\documentclass[conference]{IEEEtran}
\IEEEoverridecommandlockouts
\usepackage{cite}
\usepackage{amsmath,amssymb,amsfonts}
\usepackage{algorithmic}
\usepackage{graphicx}
\usepackage{textcomp}
\usepackage{subfigure} 
\usepackage{setspace}
\usepackage{enumerate}
\usepackage{booktabs}
\usepackage{multirow}
\usepackage{float}
\usepackage{caption}
\usepackage{makecell}
\usepackage{xcolor}
\usepackage{array}
\usepackage{colortbl}
\definecolor{mygray}{gray}{.9}

\usepackage{etoolbox}
\usepackage{balance}
\makeatletter
\patchcmd{\@makecaption}
  {\scshape}
  {}
  {}
  {}
\makeatletter
\patchcmd{\@makecaption}
  {\\}
  {.\ }
  {}
  {}
\makeatother
\def\tablename{Table}

\def\BibTeX{{\rm B\kern-.05em{\sc i\kern-.025em b}\kern-.08em
    T\kern-.1667em\lower.7ex\hbox{E}\kern-.125emX}}
\begin{document}

\title{Foresight Social-aware Reinforcement Learning for Robot Navigation 
\thanks{This work was supported by the China Scholarship Council (CSC).}
}

\author{\IEEEauthorblockN{Yanying Zhou}
\IEEEauthorblockA{\textit{Institute for Numerical Simulation} \\
\textit{University of Bonn}\\
Bonn, Germany \\
zhou@ins.uni-bonn.de}
\and
\IEEEauthorblockN{Shijie Li}
\IEEEauthorblockA{\textit{Institute for Computer Sciences} \\
\textit{University of Bonn}\\
Bonn, Germany \\
lsj@uni-bonn.de}
\and
\IEEEauthorblockN{Jochen Garcke}
\IEEEauthorblockA{\textit{Institute for Numerical Simulation} \\
\textit{University of Bonn}\\
\textit{and Fraunhofer SCAI}\\
Bonn, Germany \\
garcke@ins.uni-bonn.de}

}

\maketitle

\begin{abstract}
When robots handle navigation tasks while avoiding collisions, they perform in crowded and complex environments not as good as in stable and homogeneous environments. This often results in a low success rate and poor efficiency. 
Therefore, we propose a novel Foresight Social-aware Reinforcement Learning (FSRL) framework for mobile robots to achieve collision-free navigation. 
Compared to previous learning-based methods, our approach is foresighted. 
It not only considers the current human-robot interaction to avoid an immediate collision, but also estimates upcoming social interactions to still keep distance in the future.
Furthermore, an efficiency constraint is introduced in our approach that significantly reduces navigation time.
Comparative experiments are performed to verify the effectiveness and efficiency of our proposed method under more realistic and challenging simulated environments.
\end{abstract}


\section{Introduction}

With the uptake of artificial intelligence,  robots play a more and more important role in human life, for example as delivery or service robots.
Instead of working in a restricted space, these robots share their working space with humans and have frequent interactions, while they are expected to arrive at their destinations without a negative impact on people.

As robot navigation can be formulated as a reinforcement learning (RL) task, recently, more and more works apply RL algorithms to utilize the strong decision-making ability of RL \cite{Sutton2018}.
Meanwhile, deep learning, with its excellent representation ability, has achieved great success in many areas, like computer vision and natural language processing.
Hence, some works \cite{8593871,8794134,7989037,8202312,9197148,liu2021decentralized} combine deep learning and RL for social-aware robot navigation.
In these approaches, an effective policy is learned that implicitly models the complex interactions and cooperation of agents.
Specifically, a deep neural network is adopted to approximate the value function and the optimal action is then chosen based on this value function.
Although learning-based methods outperform non-learning methods, these methods are still limited when mobile robots navigate in more realistic complex environments. 

\begin{figure}[htbp!]
\setlength{\belowcaptionskip}{-6mm}
\centering
\includegraphics[width=0.7\linewidth, height=0.65\linewidth]{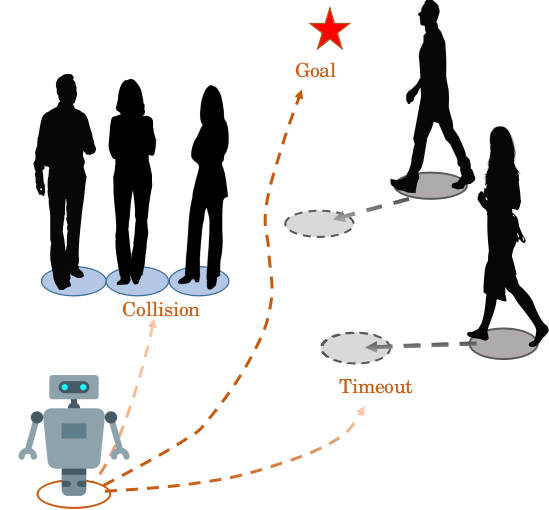}
\caption{ 
In a realistic scenario, both standing (blue) and dynamic (grey) objects exist. 
Unlike oversimplified scenarios in previous methods where only dynamic objects exist, traps or blind spots will form more frequently and not disappear over time. 
These will trap the robot in the crowd more easily.
}
\label{fig:static}
\end{figure} 

A main issue in previous robot navigation works is that they oversimplify the environment by only considering the avoidance of dynamic objects.
In this scenario, the robot usually learns a simple policy that achieves the goal by bypassing all dynamic objects.
First, it is problematic that these previous methods can behave shortsighted. 
They make the decision only according to the current interactions and ignore possible future situations, which limits their navigation ability in complex crowds.
Additionally, note that previous methods only implicitly penalize inefficient policies.
Hence, these approaches often lack the capability to navigate through complex crowds and are in particular prone to bypass them inefficiently by taking excessive detours to avoid collisions.
In more realistic scenarios where standing objects exist, these problems deteriorate further, as illustrated in Fig. \ref{fig:static}.

\begin{figure}[htbp!]
\centering
\includegraphics[width=0.7\linewidth,height=0.7\linewidth]{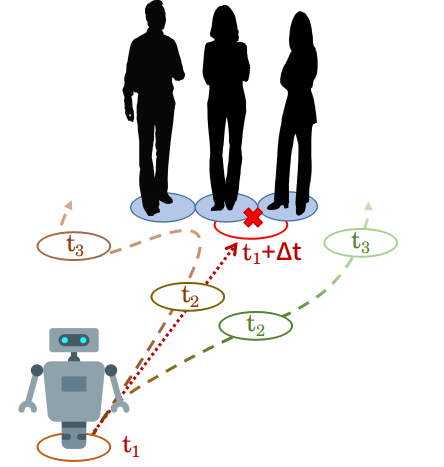}
\caption{Previous methods (brown) only detect the collision at time $t_2$ hence turn around sharply to avoid a collision, which is shortsighted. In contrast, our method (green) can forecast potential collision at time $t_1$ within $\Delta t$ (red) hence can take action in advance to avoid collision smoothly, which is foresighted. This is more important for nonholonomic robots as they cannot get out of traps easily.}
\label{fig:foresignt}
\end{figure}

Another problem in many existing works is the assumption that the robots have holonomic kinematics.
It eases the robot navigation task as the robots can freely move in any direction at any timestep.
Holonomic robots can perform large-angle rotation actions to access their destination easily with non-smooth navigation paths.
However, most robots in daily life have nonholonomic kinematics\cite{gao2021hybrid}, such as service and delivery robots.
It requires that the robots can only move smoothly, which intensifies the above problems.
For example, for trapped non-holonomic robots, it is hard to get out of their trap.



Based on the above limitations, we present a novel deep reinforcement learning (DRL) approach for robot navigation with collision avoidance. 
We enrich our RL method with the ability to forecast future potential interactions based on the current states.
In other words, the robots are more foresighted, which is shown in Fig. \ref{fig:foresignt}.
The robot can make a decision combining both current and predicted interactions, thereby reacting earlier in advance of potential collisions.
Hence, no matter the complexity of environment, it is feasible for the robot to move smoothly between successive timesteps.
In addition, our approach explicitly takes standing objects into account. 
Specifically, unlike previous methods that ignore the different kinematics between objects, we apply different restrictions on obstacles according to their movement states.
Our method can detect future potential unsafety and take corresponding actions in advance, which can be seen in Fig. \ref{fig:foresignt}.
Moreover, our method employs a constraint on the navigation time, which enables the learning of a more efficient strategy.


As our method can forecast future collisions (foresighted), we name it as Foresight Social-Aware Reinforcement Learning (FSRL) algorithm.
In summary, our contributions are:
\begin{itemize}
    \item Our method can forecast potential collisions in the future and can take actions in advance to avoid these collisions, which improves the quality of navigation.
    \item Our method explicitly uses a constraint on the navigation time, hence the learned strategy is more efficient.
    \item We perform experiments in complex environments with dynamic and standing humans.
    \item Together, we introduce a foresighted navigation method that achieves the best performance in different environments, which proves its effectiveness and robustness.
\end{itemize}

\section{RELATED WORK}

Robot navigation in pedestrian-rich environments is complicated since human behavior is diverse and stochastic depending on human intent and social interactions.
Earlier works on robot navigation used \textit{reaction-based methods} \cite{6698863,ferrer2017robot,truong2017toward,10.1007/978-3-642-19457-3_1,4543489,trautman2010unfreezing} that choose the optimal action according to one-step interaction rules. 
Social Force Model \cite{6698863,ferrer2017robot,truong2017toward} models the interactions as 'forces', such as attractive forces and repulsive forces, to drive the agent to reach the goal while avoiding obstacles.
Optimal Reciprocal Collision Avoidance (ORCA) \cite{10.1007/978-3-642-19457-3_1} and Reciprocal Velocity Obstacle (RVO) \cite{4543489} achieve collision-free navigation under reciprocal assumption. 
Interacting Gaussian Process (IGP) \cite{trautman2010unfreezing} presents an interaction potential function to couple independent Gaussian Processes.
However, these methods are limited by their hand-crafted social functions that can only capture simple interactions and are hard to be extended to crowded situations.
Also, they are prone to lack foresight and have unnatural behaviors \cite{liu2021decentralized}.

\textit{Trajectory-based methods} \cite{li2020socially,aoude2013probabilistically,vemula2017modeling} plan a proper path for robots by predicting other agents' trajectories from the large-scale datasets.
\cite{aoude2013probabilistically} combines Gaussian Processes with Rapidly-exploring Random Trees to generate probabilistic safe paths.
\cite{vemula2017modeling} presents a spatially local interaction function to predict the joint human motion.
Although trajectory-based methods are long-sighted, they are computationally expensive and time-consuming by explicitly accounting evolution of joint paths, especially when there are large teams.
In addition, since robots take overly conservative strategies, robots have less planable navigation space and thus are more likely to face the freezing robot problem \cite{7989037}.  

Compared to these non-learning-based methods, \textit{learning-based methods} are more computationally efficient and can perform much better.
DRL methods \cite{8593871,8794134,7989037,8202312,9197148,liu2021decentralized} explore a computationally efficient interaction rule by pre-training a value function.
The optimal actions can then be obtained based on the currently observed states.
The robot can thus avoid collisions while planning the trajectory.
In SA-CADRL \cite{8202312} the optimal action is chosen through a maxmin operation.
SARL\cite{8794134} jointly models the human-robot interactions as well as human-human interactions based on CADRL \cite{7989037}.
\cite{9197148} proposes a framework by combining ego-safety and social-safety in mapless navigation.

\begin{figure*}[htbp]
\setlength{\belowcaptionskip}{-6mm}
\centering
\subfigure[Effective range $r_{e}$]{
\begin{minipage}[t]{0.33\linewidth}
\centering
\includegraphics[width=0.85\linewidth]{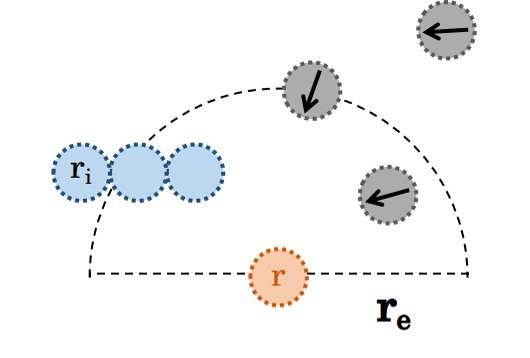}
\end{minipage}%
}%
\subfigure[Static situation]{
\begin{minipage}[t]{0.33\linewidth}
\centering
\includegraphics[width=0.65\linewidth,height=4cm]{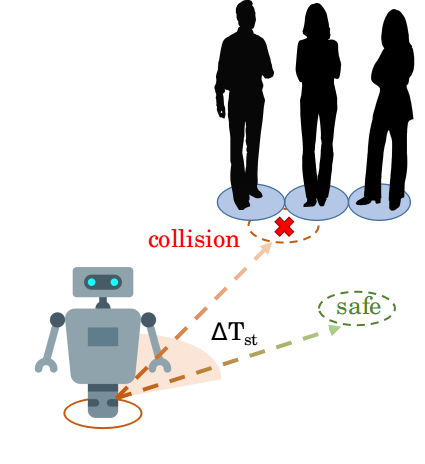}
\end{minipage}%
}%
\subfigure[Dynamic situation]{
\begin{minipage}[t]{0.33\linewidth}
\centering
\includegraphics[width=0.85\linewidth,height=4cm]{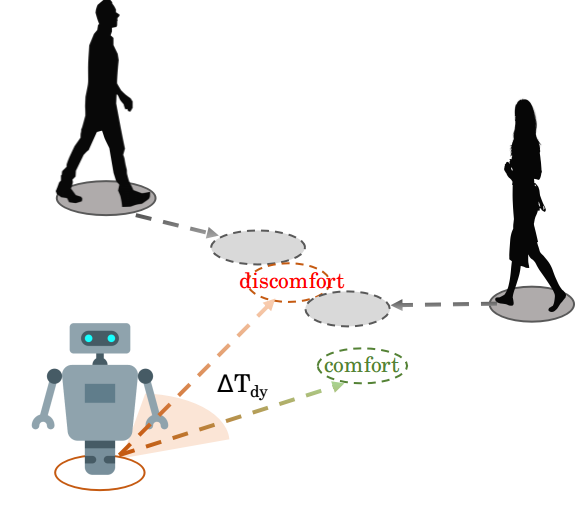}
\end{minipage}
}%
\centering
\caption{
(a) An illustration of the effective range $r_{e}$ of the robot. The blue and grey circles respectively represent the standing and dynamic humans with radius ${r}^{i}$, where arrows indicate the moving direction. (b) By detecting potential collisions with standing objects in $\Delta T_{st}$, the robot can take action in advance to avoid collisions. This is achieved by the reward component $R_{st}$.
(c) According to the comfortable distance $r_c$ within $\Delta T_{dy}$, the robot chooses to detour the crowd instead of going through it, which would annoy humans. This is achieved by the reward component $R_{dy}$.}
\label{fig:RRR}
\end{figure*}

However, current methods do not perform well in realistic environments.
It is because they oversimplify the environment and robotics kinematics in robot navigation which are far from realistic scenarios.
This leads the robots are usually shortsighted in crowded or movement restricted environments, while lacking the ability to be aware of potential collisions.
In contrast, our method estimates not only the current interactions but also predicts future potential situations.
Hence, it can navigate through complex crowds smoothly and balance safety and efficiency.
Furthermore, our method is more efficient by explicitly introducing constraints on the navigation time.

\section{APPROACH}

\subsection{Problem Formulation}\label{AA}

We formulate social robot navigation in 2D space as a sequential decision making problem in a RL framework, which is a Partially Observable Markov Decision Process (POMDP).
At each time step $t$,  for each agent (robot or human), we denote its state and action as $\mathbf{s}_t$ and $\mathbf{a}_t$, respectively. 
The state ${\mathbf{s}}_t$ can be divided into observed and hidden parts, that is ${\mathbf{s}_t}=[\Tilde{\mathbf{s}}_t,~\hat{\mathbf{s}}_t]$.
Here, the observable part is $\Tilde{\mathbf{s}}_t = [{p}_x, {p}_y, {v}_x, {v}_y, {r}] \in \mathbb{R}^5$, where $\mathbf{{p}}=$ $[{p}_x,{p}_y]$, $\mathbf{{v}}=[{v}_x,{v}_y]$ and ${r}$ are the agent's position, velocity, and radius, respectively.
The hidden part is $\hat{\mathbf{s}}_t=[{g}_x, {g}_y, v_{pref}, {\theta}] \in \mathbb{R}^4$, where $\mathbf{{p}}_g=[{g}_x, {g}_y]$, $v_{pref}$ and ${\theta}$ are the goal position, preferred speed, and orientation angle, respectively.
We denote all humans' observed state as $\Tilde{\mathbf{s}}_t^H = [\Tilde{\mathbf{s}}_t^1,\Tilde{\mathbf{s}}_t^2,\dots,\Tilde{\mathbf{s}}_t^n]$ at time step $t$.
For simplicity, we use $\mathbf{s}_t$ to denote the robot state.
Thus, the joint state observed by the robot is defined as $\mathbf{S}_t=[\mathbf{s}_t,\Tilde{\mathbf{s}}_t^H]$.

Considering that the robot has unicycle nonholonomic kinematics, the action $\mathbf{a}_t$ can be denoted by linear and angular velocities, $\mathbf{a}_t=[v_t,\omega_t]\in\mathbb{R}^2$. 
Following the RL methodology~\cite{Sutton2018}, the robot will receive a reward $R_{t}$ when performing an action to judge the quality of its decision. 
The goal is to find an optimal policy $\pi^{\ast}: \mathbf{S}_t \mapsto \mathbf{a}_t$ that assigns each state an action to maximize the expected future cumulative reward of the actions taken till the goal is reached.
A value network is designed to encode an estimate of the optimal value function:
\begin{equation}
V^{\ast}(\mathbf{S}_t) = \mathbb{E} \left[ {\sum_{t'=t}^{T}{\gamma^{t'\cdot v_{pref}}R\left(\mathbf{S}_{t'},\pi^{\ast}(\mathbf{S}_{t'})\right)}} \right],
\end{equation}
where $\gamma \in (0,1)$ is a discount factor, and $R( {\mathbf{S}_t,\mathbf{a}_t})$ is the received reward at time $t$.
The optimal policy $\pi^{\ast}{({\mathbf{S}_t})}$ can be determined after computing the value function $V^{\ast}(\mathbf{S}_t)$:
\begin{equation}
\begin{split}
\pi^{\ast}&{({\mathbf{S}_t})}  = \mathop{\arg\!\max}_{\mathbf{a}_t} R (\mathbf{S}_t,\mathbf{a}_t) +  \\
& \gamma^{\Delta t \cdot v_{pref}}\!\!\int_{\mathbf{S}_{t+\Delta t}}\!\!{P(\mathbf{S}_t,\mathbf{a}_t,\mathbf{S}_{t+\Delta t})V^{\ast}(\mathbf{S}_{t+\Delta t}) \, d\mathbf{S}_{t+\Delta t}},
\end{split}
\end{equation}
where $\Delta t$ is the time step. $P(\mathbf{S}_t,\mathbf{a}_t,\mathbf{S}_{t+\Delta t})$ denotes the transition probability between $t$ and $t+\Delta t$. 

\subsection{Parametrization}

We set the robot as the center of the coordinate system with its first-person perspective as $x$-axis pointing towards the goal.
Based on this transformation, the states of the robot and humans can be represented by:
\begin{equation}
    \begin{split}
        & s_t = [d_g, v_{pref}, v_{t}, \omega_{t}, r],\\
        & \Tilde{s}_t^{{i}} = [{p}_x^i, {p}_y^i, {v}_x^i, {v}_y^i, {r}^{i}, {d}^i, {r}^{i}+r],
    \end{split}
\end{equation}
where $d_g = \lVert{\mathbf{p}-\mathbf{p}_g}\rVert_2$ and ${d}^i = \lVert{\mathbf{p}-\mathbf{{p}}^i}\rVert_2$ respectively represent the distance between the robot and its goal and the distance between the robot and the human $i$.

\begin{figure*}[htbp]
\setlength{\abovecaptionskip}{-0.5mm}
\setlength{\belowcaptionskip}{-6mm}
\centering
\subfigure[Env.1]{
\begin{minipage}[t]{0.33\linewidth}
\centering
\includegraphics[width=0.7\linewidth]{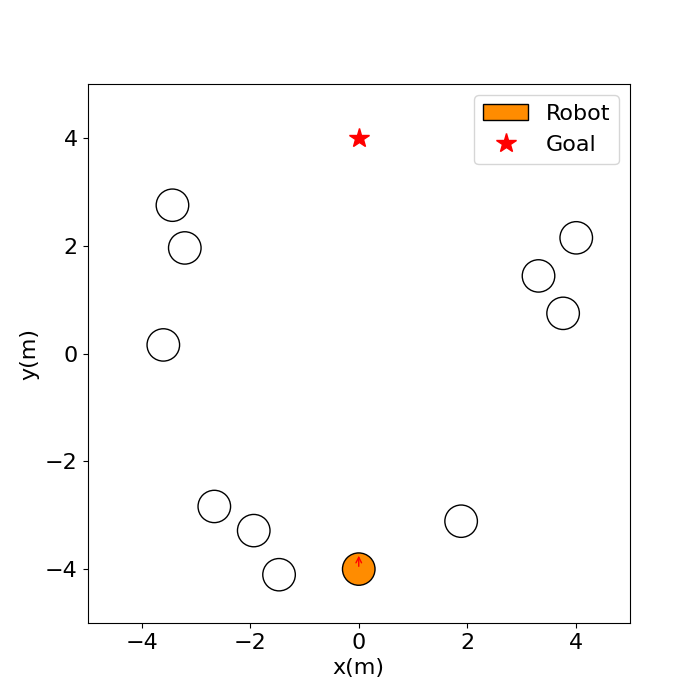}
\end{minipage}%
}%
\subfigure[Env.2]{
\begin{minipage}[t]{0.33\linewidth}
\centering
\includegraphics[width=0.7\linewidth]{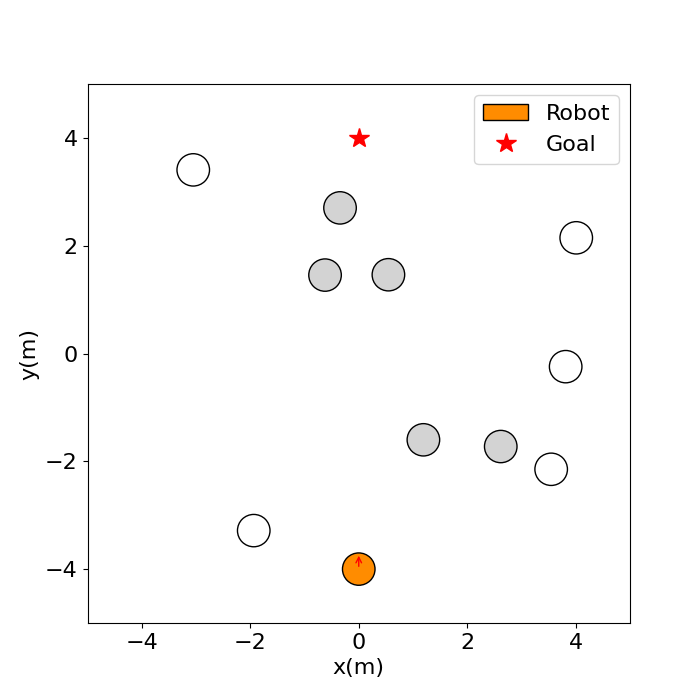}
\end{minipage}%
}%
\subfigure[Env.3]{
\begin{minipage}[t]{0.33\linewidth}
\centering
\includegraphics[width=0.7\linewidth]{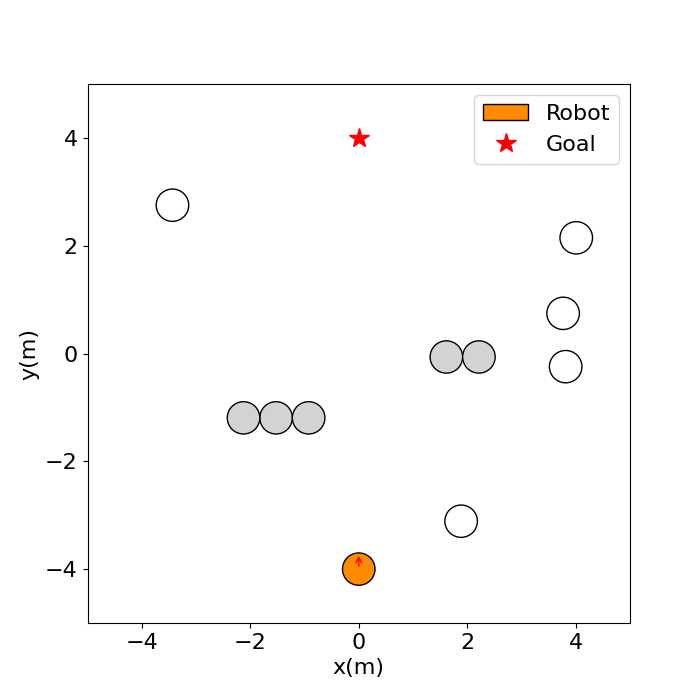}
\end{minipage}
}%
\centering
\caption{Simulation environments. Hollow circles are dynamic objects, whereas solid grey circles are static standing agents.  
(a) Env.1 has 10 dynamic objects.
(b) Env.2 has 5 random standing objects and others are dynamic.
(c) Env.3 has two group barriers of 2 and 3 standing objects and others are also dynamic.
Compared to the previous environment, robots are more easily trapped and hard to detour in our latter two environments.
}
\label{fig:simulate_env}
\end{figure*}

\subsection{Reward Function}

In previous methods \cite{8794134,7989037}, the decision at each time step is only determined using the current state, which is formulated by the reward function $R_c(\mathbf{S}_{t},\mathbf{a}_t)$:
\begin{equation}
    R_c(\mathbf{S}_{t},\mathbf{a}_t) = 
    \begin{cases}
    -0.25 & \mbox{if $d_t < 0$} \\
    0.25(-0.1+d_t/2) & \mbox{else if $d_t < r_{c}$} \\
    1 & \mbox{else if $p = p_g$} \\
    0 & \mbox{otherwise},
    \end{cases}
\end{equation}
where $d_t = min{\left\{{d}^{i}-r-{r}^i\right\}}$ denotes the closest distance from the robot to the other humans at time $t$. The comfortable distance $r_{c} = 0.2m$ is defined as the minimum comfort distance between the humans and the robot. 

One reason for the poor performance of previous methods in complex environments is that they are easily trapped, mainly since they cannot detect potential collisions in the future.
This deteriorates further when more realistic nonholonomic kinematics are adopted as sharp movements are then forbidden.
Thus, we propose a foresighted method, which means to take actions in advance.
Since in nonholonomic kinematics the robot's rotation angle is limited, its movement in a short time period are smooth and can be seen as linear.
Future situations then can be estimated from the current state.
Based on this, we introduce a foresight penalty $R_f$ that penalizes potential collisions in the future.
Furthermore, in a realistic environment, both dynamic and standing objects exist, which we in contrast to previous methods distinguish.
Our method classifies them based on their velocities and treats them differently. 
Hence an improved strategy can be learned. 

The foresight penalty $R_f$ is now formulated as:
\begin{equation}
    R_f(\mathbf{S}_{t},\mathbf{a}_t) = R_{dy}(s_{t}^{jn},\mathbf{a}_t)+R_{st}(s_{t}^{jn},\mathbf{a}_t),
\end{equation} 
where $R_{dy}$ and $R_{st}$ are responsible for dynamic and standing objects, respectively, and are introduced in the following.

As far objects have little influence on the current decisions, we only take the objects within the effective range $r_e=2m$ into account, shown in Fig. \ref{fig:RRR}(a).
For standing objects in $r_e$, the robot can only choose to bypass them as they will not disappear over time.
If the robot detects that there are potential collisions with standing objects in $\Delta T_{st}=2s$ according to the current decision, a penalty is applied.
This penalty is proportional to the number of potential collisions with standing objects, which reveals a crowding level, and is formulated as:
\begin{equation}
    R_{st}(\mathbf{S}_{t},\mathbf{a}_t) = - \alpha \ast \frac{N_{col}}{N_{static}},
\end{equation}
where $\alpha = 0.15$. $N_{col}$ is the number of detected potential collisions between standing objects (in $r_e$) and the robot in $\Delta T_{st}$. $N_{static}$ is the number of standing humans that are in the effective range $r_e$ of the robot.  
By this reward penalty, potential collision are encouraged to be avoided, while as an effect of the foresight built into this penalty the robot also shall take actions in advance to avoid sharp movements, hence making the navigation more smooth, see Fig. \ref{fig:RRR}(b).

As for dynamic objects, they easily gather into clusters and thereby cause collisions. 
Different from the above scenario, these clusters will disappear as time goes by.
Thus the robot needs to determine whether to wait and then go through or to bypass clusters.
Further, in a real-world application, the robot and humans are expected to not influence each other.
Therefore, apart from navigating to the goal successfully, we also care about the perceived quality of navigation.
Since dynamic humans would move to achieve a comfortable distance, an additional penalty will be applied when the robot disturbs the people (their distance is less than the comfortable distance).
This can be formulated as:
\begin{equation}
    R_{dy}(\mathbf{S}_{t},\mathbf{a}_t) = \beta \ast (d_{\Delta T_{dy}} - r_c),
\end{equation}
where $\beta = 0.5$, $r_c=0.2m$. $d_{\Delta T_{dy}}$ is the minimum distance between the dynamic humans and the robot after $\Delta T_{dy}=1s$,
which is half of $\Delta T_{st}$ in $R_{st}$ as dynamic objects move and 
thus this time window needs to be shorter.
By this reward penalty, the robot shall avoid aggressive navigation, in other words, it avoids disturbing people frequently, see Fig. \ref{fig:RRR}(c).

Apart from the success rate of navigation, we also explicitly model efficiency in the reward, which is often neglected in previous methods.
To achieve this, we add a constraint on the navigation time  and encourage efficient strategies by: 
\begin{equation}
    R_t(\mathbf{S}_{t},\mathbf{a}_t) = 
    \begin{cases}
    - 0.1 \ast \frac{t}{t_{limit}} & \mbox{if $p = p_g$} \\
    - 0.2 & \mbox{else if $t >= t_{limit}$}, \\
    \end{cases}   
\end{equation}
where $t_{limit}$ is the maximum navigation time, which is set as $t_{limit}=25s$ in our experiments. 
By adding this term, detours will be discouraged.

Together, the whole reward function $R$ is defined as:
\begin{equation}
    R(\mathbf{S}_{t},\mathbf{a}_t) = R_c(\mathbf{S}_{t},\mathbf{a}_t) + R_f(\mathbf{S}_{t},\mathbf{a}_t) + R_t(\mathbf{S}_{t},\mathbf{a}_t).
\end{equation}

\begin{figure*}[htbp]
\setlength{\abovecaptionskip}{-0.5mm}   
\setlength{\belowcaptionskip}{-3mm}
\centering
\subfigure[Invisible Setting]
{
\begin{minipage}[b]{0.17\linewidth}
\centering
\includegraphics[width=\linewidth]{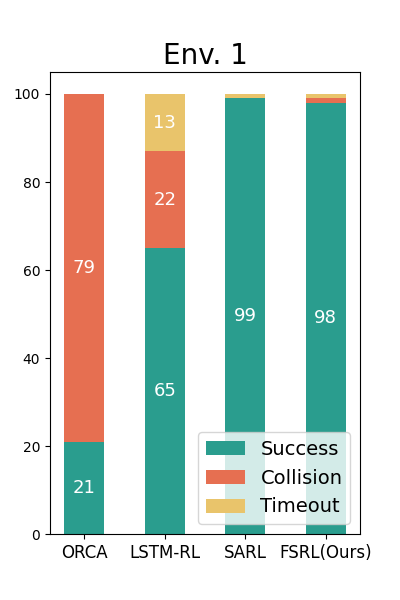}
\end{minipage}%
\begin{minipage}[b]{0.17\linewidth}
\centering
\includegraphics[width=\linewidth]{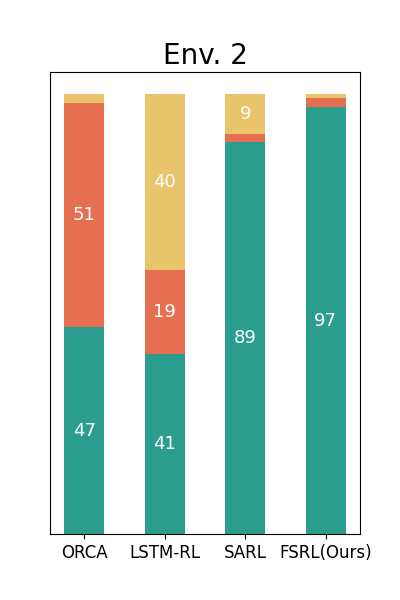}
\end{minipage}%
\begin{minipage}[b]{0.17\linewidth}
\centering
\includegraphics[width=\linewidth]{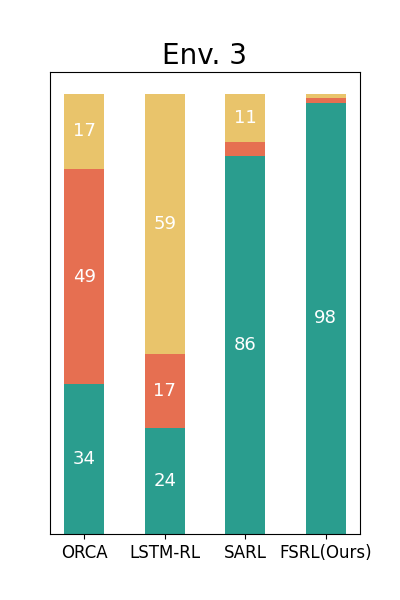}
\end{minipage}%
}%
\subfigure[Visible Setting]
{
\begin{minipage}[b]{0.17\linewidth}
\centering
\includegraphics[width=0.99\linewidth]{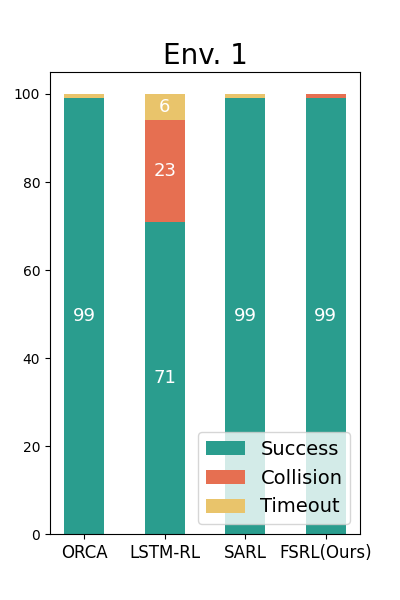}
\end{minipage}%
\begin{minipage}[b]{0.17\linewidth}
\centering
\includegraphics[width=0.99\linewidth]{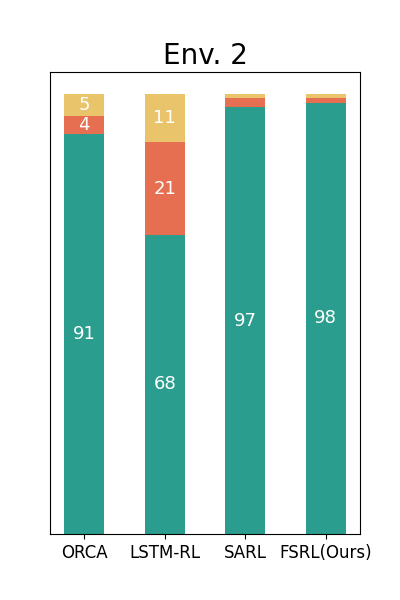}
\end{minipage}%
\begin{minipage}[b]{0.17\linewidth}
\centering
\includegraphics[width=0.99\linewidth]{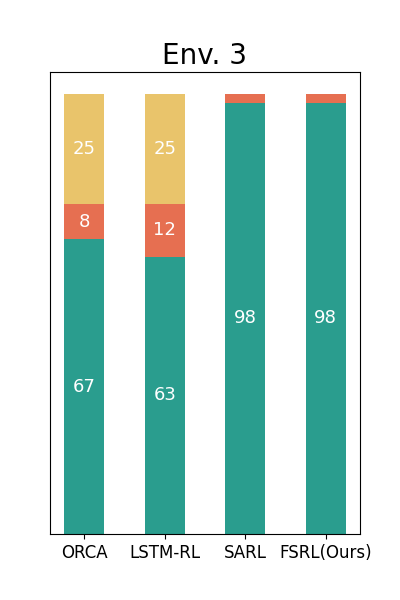}
\end{minipage}%
}%
\caption{Quantitative evaluation among ORCA\cite{10.1007/978-3-642-19457-3_1}, LSTM-RL\cite{8593871}, SARL \cite{8794134} and FSRL(Ours) methods in three environments: (a) The three pictures on the left are with invisible setting; (b) The three pictures on the right are with visible setting.}
\label{fig:sr}
\end{figure*}


\subsection{Training the Policy}

We train the policy using the proposed reward via a deep V-learning algorithm \cite{8794134}.
The training process can be divided into two stages. 
First, the policy is initialized via imitation learning by collecting 3000 episodes demonstrations based on the ORCA \cite{10.1007/978-3-642-19457-3_1} policy, which is trained using the Adam optimizer\cite{kingma2017adam} with 50 epochs and a learning rate of 0.01.  
Then, the policy is refined based on the proposed reward functions with the learning rate set to 0.0001. 
In addition, the discount factor $\gamma$ is 0.9. 
In a greedy fashion, the exploration rate decays linearly from 0.5 to 0.1 in the first 4000 episodes and remains at 0.1 for another 6000 episodes. 
We assume that the robot has nonholonomic kinematics, which is a constraint on the rotation angle. 
The robot's velocity is exponentially discretized into 5 speeds between (0, $V_{pref}$] and 10 headings spaced between $[-\pi/8,\pi/8]$. 


\section{EXPERIMENTS}


\subsection{Simulation Setup}

We build the simulation environment using Gym\cite{brockman2016openai} and RVO\cite{4543489} as in \cite{8794134}.
There is one robot and $10$ dynamic/standing humans in the environment.
Our goal is to learn an optimal strategy to make the robot navigate to the destination efficiently.
In each experiment, the start and goal position of the robot are set to $(0, -4)$ and $(0, 4)$, respectively.
Previous works adopt a circle-crossing where all humans are dynamic and randomly placed on a circle with their destinations on the opposite side.
Further, all humans act following the ORCA policy\cite{10.1007/978-3-642-19457-3_1}.
As mentioned, this scenario is oversimplified and objects can be easily bypassed, which is far from a realistic environment.
Hence, one can not demonstrate the navigation ability of algorithms comprehensively.

To better show the navigation ability, we test all methods on \textit{three increasingly challenging environments}, shown in Fig.\ref{fig:simulate_env}.
\begin{itemize}
    \item \textbf{Env. 1} is the same as in previous works where only dynamic agents exist.
    \item \textbf{Env. 2} contains 5 dynamic and 5 standing agents. The positions of the standing agents are distributed randomly in the whole environment. We can see that traps and blind areas are more easily formed in this environment.
    \item \textbf{Env. 3} is an extreme version of Env. 2. The only difference is that the standing agents form two fixed traps and blind areas that will not disappear over time. 
\end{itemize}

In addition, we perform experiments in two settings: 
one sets the robot \textbf{invisible} to humans, which means that humans do not react to the robot's behavior, such as giving way;
the other is the \textbf{visible} setting, where the mutual human-robot interactions increase the uncertainty of environments. 
In either setting, the robot needs a more foresighted policy. 


We compare our method with three state-of-the-art methods: ORCA\cite{10.1007/978-3-642-19457-3_1} is the typical reaction-based method and LSTM-RL\cite{8593871}, SARL \cite{8794134} are DRL methods. 
Specifically, for the three environments and the two settings for each environment (6 environments in total), we train the unicycle rotation-constrained robot with all four algorithms separately and then obtain the corresponding test results. 
To reduce the influence of randomness, we conduct 3 experiments for each method on each environment and report the average performance. 

\subsection{Metrics}

For each environment, we obtain the quantitative evaluation after 500 random test episodes based on the trained models.   

There are five metrics used in the quantitative evaluation: 
\begin{enumerate}[(1)]
    \item "Succ.(\%)": the rate of success cases where the robot reaches its goal without a collision in the maximum allowed time $t_{limit}$. 
    \item "Coll.(\%)": the rate of collision cases where the robot collides with other agents. 
    \item "Timeout(\%)": the rate of timeout cases where the robot neither collides with others nor reaches its goal in the maximum allowed time $t_{limit}$.
    \item "Nav. Time": average navigation time of the above success cases in seconds.
    \item "Disc. (\%)": average frequency of duration that the robot has distances to any other agents less than comfortable distance (0.2m).
\end{enumerate}






\subsection{Comparison with State-of-the-art Methods}

We show the comparison of performance between FSRL (ours) and state-of-the-art methods in Fig. \ref{fig:sr} and Tab. \ref{tab1}, \ref{tab2} under invisible and visible settings.
We can easily find that FSRL achieves the best performance on almost all scenarios. 

Fig. \ref{fig:sr} shows the success, collision and timeout rates.
No matter if in the invisible or visible setting, FSRL achieves the highest success rates and the lowest rates of collision and timeout.
Since learning-based methods, such as LSTM-RL or SARL, choose the optimal action only based on the current states, they are usually short-sighted.
In particular, a nonholonomic robot, i.e. with limited rotations, cannot rotate at a wide angle when meeting obstacles, so it needs longer navigation time, easier gets trapped, or even fails.
As shown in Fig. \ref{fig:sr}, there also are more timeouts or collisions.
As expected, FSRL performs the best as it is foresighted and not only considers the current state but also estimates future situations.
The nonholonomic robot thus can take proper actions in advance to move smoothly and safely.

\begin{table}[!htbp]
    \renewcommand\arraystretch{1.5}
	\centering
	\captionsetup{font={normalsize}}
 	\caption{"Nav. Time" quantitative results in 3 environments}
    \resizebox{\linewidth}{!}{
    \begin{tabular}{c|c|c|c|c|c|c|c|c}
    \hline
    \multirow{2}*{Methods}& \multicolumn{4}{c|}{Nav Time (invisible)}& \multicolumn{4}{c}{Nav Time (visible)} \\
    \cline{2-9}
    &Env.1&Env.2&Env.3&AVG&Env.1&Env.2&Env.3&AVG\\
    \hline
    ORCA\cite{10.1007/978-3-642-19457-3_1}&12.49&11.98&12.04&12.14&11.99&10.97&10.74&11.23\\
    LSTM-RL\cite{8593871}&15.52&14.12&12.81&14.15&11.01&10.96&12.32&11.29\\
    SARL\cite{8794134}&12.25&11.39&\cellcolor{mygray}11.15&11.60&\cellcolor{mygray}9.85&10.33&\cellcolor{mygray}10.01&\cellcolor{mygray}10.06\\
    \cellcolor{mygray}FSRL(\textit{ours})&\cellcolor{mygray}10.90&\cellcolor{mygray}10.92&11.69&\cellcolor{mygray}11.04&11.08&\cellcolor{mygray}10.32&11.60&11.00\\
    \hline
    \end{tabular}}
	\label{tab1}
\end{table}

\begin{table}[!htbp]
    \renewcommand\arraystretch{1.5}
	\centering
	\captionsetup{font={normalsize}}
 	\caption{"Disc. (\%)" quantitative results in 3 environments.}
    \resizebox{\linewidth}{!}{
    \begin{tabular}{c|c|c|c|c|c|c|c|c}
    \hline
    \multirow{2}*{Methods}& \multicolumn{4}{c|}{Disc(\%) (invisible)}& \multicolumn{4}{c}{Disc(\%) (visible)}\\
    \cline{2-9}
    &Env.1&Env.2&Env.3&AVG&Env.1&Env.2&Env.3&AVG\\
    \hline
    ORCA\cite{10.1007/978-3-642-19457-3_1}&0.39&0.38&0.43&0.40& 0.41&0.41&0.45&0.42\\
    LSTM-RL\cite{8593871}&0.07&0.28&0.16&0.17&0.11&0.35&0.39&0.20\\
    SARL\cite{8794134}&\cellcolor{mygray}0.01&\cellcolor{mygray}0.06&0.06&\cellcolor{mygray}0.04&0.09&0.11&0.10&0.10\\
    \cellcolor{mygray}FSRL(\textit{ours})&\cellcolor{mygray}0.01&0.07&\cellcolor{mygray}0.04&\cellcolor{mygray}0.04&\cellcolor{mygray}0.06&\cellcolor{mygray}0.08&\cellcolor{mygray}0.07&\cellcolor{mygray}0.07\\
    \hline
    \end{tabular}}
	\label{tab2}
\end{table}

In the \textbf{invisible} setting, as anticipated, the ORCA method seriously fails due to the violated reciprocal assumption. 
However, it is interesting that the ORCA robot has much better performance under Env. 2, 3 than under Env. 1.
This is because non-learning-based methods, like ORCA, are more sensitive to dynamic objects.
Hence, fewer dynamic objects will lead to better performance.

In contrast, there is a significant performance drop for learning-based methods (LSTM-RL, SARL) when Env. 1 is transformed into Env. 2, 3.
This is due to them having difficulties to learn a good representation for a complex environment where both standing and dynamic objects exist, compared to a simpler environment where either standing or dynamic objects exist.
It demonstrates that previous learning-based methods are usually suitable for simple scenarios and do not perform well in a more complex environment.
In comparison, the LSTM-RL method is more conservative, which leads to timeout cases increasing sharply and also to longer navigation times in success cases (Tab. \ref{tab1}) .
In contrast, FSRL has the shortest navigation time and the lowest discomfort rate.
These results demonstrate the foresightedness of FSRL, allowing for a better balance between safety and efficiency.


In the visible setting, although the robot more easily navigates through the crowd due to the aid by the give-way behaviors of humans, it needs to understand more about interactions with humans to face the increasing uncertainty. 
For example, there is an increased likelihood that humans with give-way behavior will enter the comfort zone of the robot.
In addition, due to the mutual human-robot interaction, the robot is prone to take risky behaviors to achieve better navigation.
Thus, unlike in the invisible case, the performance of all methods improves except for the discomfort rate.

For LSTM-RL and ORCA methods, they are shortsighted and even fail to take standing agents into account in Env. 2,3. 
They thus navigate unsafely as well as inefficiently, as seen by a low success rate, a longer navigation time, and a high discomfort rate.
As shown in Fig. \ref{fig:sr}, FSRL performs as well as the SARL method.
Compared with SARL, although FSRL takes longer (less than 1s on average) in terms of the navigation time, it decreases the discomfort rate by 3\%  by moving the robot more safely (Tab. \ref{tab2}).

We can see the FSRL metrics in the visible cases are as well as in the invisible cases.
In an overall comparison to previous methods, FSRL is more effective, efficient, and robust.

\subsection{Ablation Study}

\begin{figure}[t]
\centering
\includegraphics[width=0.8\linewidth]{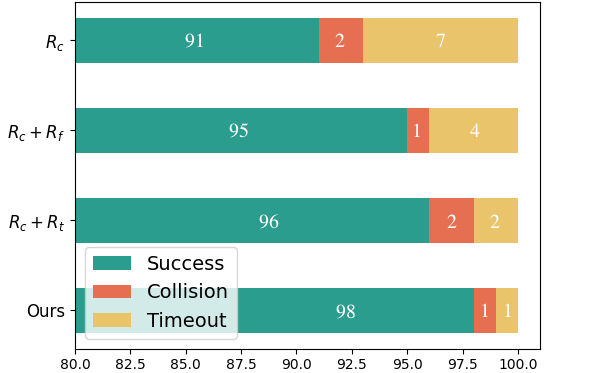}
\caption{ 
Average quantitative results under three environments of ablation experiments. (Invisible setting)
}
\label{fig:aver-quan}
\end{figure} 

\begin{figure}[t]
\centering
\includegraphics[width=0.75\linewidth]{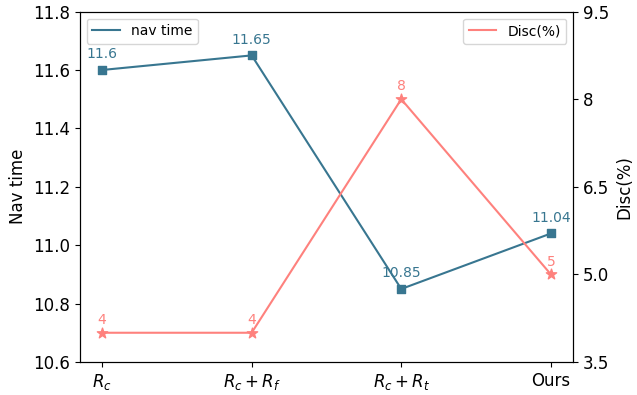}
\caption{ 
Average "Nav. Time" and "Disc($\%$)" results under three environments of ablation experiments.
(Invisible setting)}
\label{fig:line}
\end{figure} 

\begin{figure*}[t]
\setlength{\belowcaptionskip}{-6mm}
\centering
\subfigure[SARL (t = 6)]{
\begin{minipage}[t]{0.24\linewidth}
\centering
\includegraphics[width=0.9\linewidth]{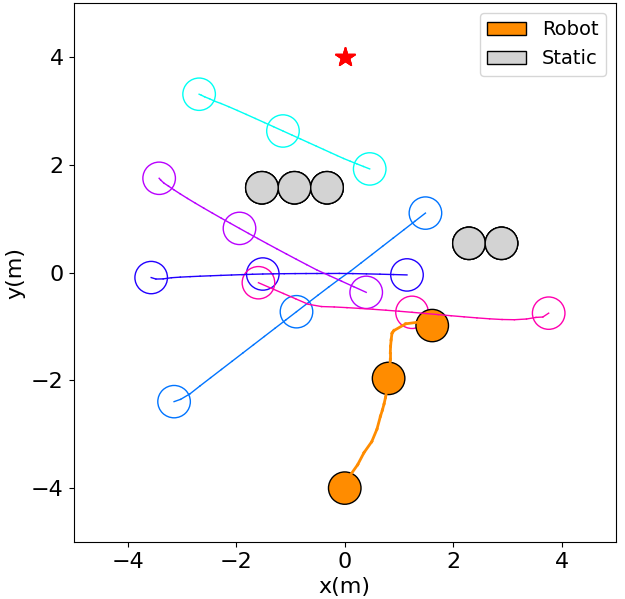}
\end{minipage}%
}%
\subfigure[SARL (t = 14.25)]{
\begin{minipage}[t]{0.24\linewidth}
\centering
\includegraphics[width=0.9\linewidth]{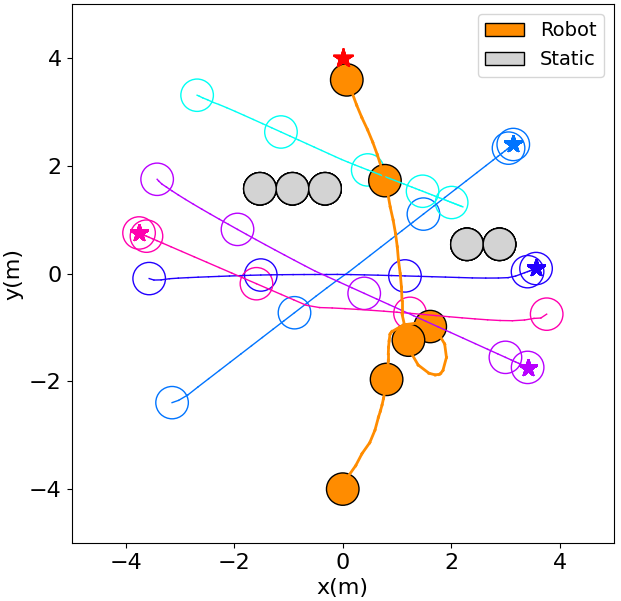}
\end{minipage}%
}%
\subfigure[FSRL (t = 6)]{
\begin{minipage}[t]{0.24\linewidth}
\centering
\includegraphics[width=0.9\linewidth]{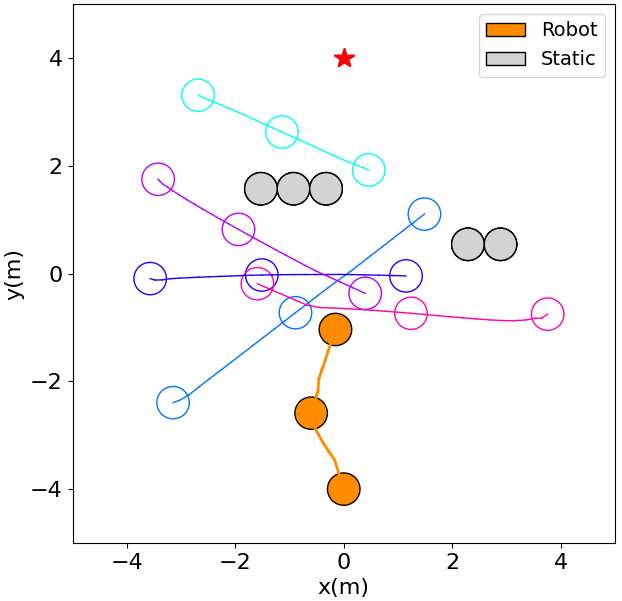}
\end{minipage}%
}
\subfigure[FSRL (t = 11.50)]{
\begin{minipage}[t]{0.24\linewidth}
\centering
\includegraphics[width=0.9\linewidth]{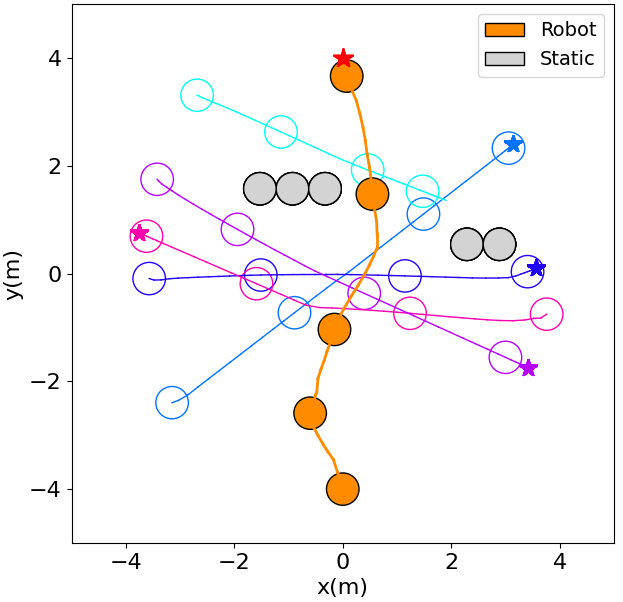}
\end{minipage}%
}%
\centering
\caption{Qualitative results. We find our method (FSRL) can make robot (orange) navigate more smoothly and efficiently in a complex environment compared to SARL\cite{8794134} where standing (grey solid) and dynamic (other colors) objects exist.
}
\label{fig:vis_refine}
\end{figure*}

We show the impact of each component of our reward $R$ in Fig. \ref{fig:aver-quan} and \ref{fig:line}, where $R_c$ is the reward that only considers the current state, Eq.(4), $R_f$ is the foresight penalty, Eq. (5), and $R_t$ is the efficiency constraint, Eq. (8).
For clarity, we take the average of the quantitative results under three environments for comparison.
In Fig. \ref{fig:aver-quan}, we can observe that both foresight penalty ($R_f$) and efficiency constraints ($R_t$) can improve the success rate (Succ.) a lot.
Furthermore, for the efficiency constraint ($R_t$), the improvement mainly comes from shortening the navigation time shown in Fig. \ref{fig:line}.
In other words, some timeout cases are avoided due to higher efficiency,
as can be seen by the reduced navigation time.
However, this improvement is at the cost of aggressive navigation,
resulting in a much higher discomfort rate (Disc.), which means the robot will disturb people intensively to save navigation time.
In contrast, the foresight penalty ($R_f$) can improve the success rate (Succ.) with less disturbance by caring more about a comfortable distance.
By combing both foresight penalty ($R_f$) and efficiency constraints ($R_t$), our method can achieve the best success rate (Succ.), while balancing efficiency and human feeling (less disturbance) well.
This demonstrates the effectiveness of our design. 

\subsection{Qualitative Evalutaion}
We show qualitative results for Env. 2 (invisible setting) in Fig. \ref{fig:vis_refine}.
At time $t=6$, SARL turns around sharply close to people.
It is because SARL makes decisions only according to the current state, it hence neglects potential collisions in the future.
Furthermore, it also results in a longer navigation time (14.25s) to arrive at the goal.
As a comparison, the navigation trajectory of FSRL is much more smooth.
This is due to FSRL's capability to forecast potential collisions in the future and to take corresponding actions in advance.
The navigation time is also shorter ($t = 11.50$), which shows the efficiency of FSRL.
Hence FSRL learns a better strategy that is more effective and efficient.

\section{CONCLUSIONS}

In this work, we propose a novel Foresight Social-Aware Reinforcement Learning (FSRL) framework for robotics navigation.
Our method can estimate potential collisions in the future and hence takes action in advance to avoid collisions.
In contrast, previous methods only make the decision according to the current state thus perform not well when the environment becomes complex.
Furthermore, an efficiency constraint is introduced in our work that reduces navigation time a lot.
We conduct experiments on three increasingly challenging environments and our method achieves the best performance regarding both effectiveness and efficiency.






{
\small
\bibliographystyle{IEEEtran}
\bibliography{ref.bib}
}

\end{document}